\documentclass{article}
\usepackage{amssymb,amsmath}
\usepackage{subcaption}
\usepackage{graphicx}
\usepackage{authblk}
\usepackage[authoryear]{natbib}

\def\Ttheta{\boldsymbol\theta}
\def\TC{\mathbf C}

\def\by{\mathbf y}

\begin{document}

\title{Automatic classification of geologic units in seismic images using partially interpreted examples.}

\author[1]{Bas Peters}
\author[2]{Justin Granek}
\author[1]{Eldad Haber}

\affil[1]{University of British Columbia, Vancouver, Canada}
\affil[2]{Computational Geosciences Inc.}
\date{January 11, 2019}
\maketitle

\begin{abstract}
Geologic interpretation of large seismic stacked or migrated seismic images can be a time-consuming task for seismic interpreters. Neural network based semantic segmentation provides fast and automatic interpretations, provided a sufficient number of example interpretations are available. Networks that map from image-to-image emerged recently as powerful tools for automatic segmentation, but standard implementations require fully interpreted examples. Generating training labels for large images manually is time consuming. We introduce a partial loss-function and labeling strategies such that networks can learn from partially interpreted seismic images. This strategy requires only a small number of annotated pixels per seismic image. Tests on seismic images and interpretation information from the Sea of Ireland show that we obtain high-quality predicted interpretations from a small number of large seismic images. The combination of a partial-loss function, a multi-resolution network that explicitly takes small and large-scale geological features into account, and new labeling strategies make neural networks a more practical tool for automatic seismic interpretation.
\end{abstract}

\section{Introduction}

Given a seismic image that is the result of stacking or migration processing in terms space and time coordinates, the next step is generating a geological interpretation. An interpretation of the seismic image includes delineating geological features, such as faults, chimneys, channels and important horizons. In this work, we focus on horizons and geological units. If two horizons are the upper and lower boundary of a specific rock type or unit, the area in between the two horizons corresponds to a single geologic entity. This is not a requirement for the methods and examples we present. Any two horizons may also contain an arbitrary stack of layers and rock types in between. We can still consider this as a single meta-unit and aim to classify it as such. 

Neural networks have a long history of learning from example seismic interpretations, to help and speed up manual interpretation. Classic works \citep{A226265, Veezhinathan1993} were, because of computational resources, limited to train on a single (window of a) time-recording, or a small portion of a seismic image at a time. Various more recent works pose geological interpretation problems as the binary classification of the central pixel of a small patch/volume of seismic data \citep{doi:10.1111/j.1365-2478.2005.00489.x, doi:10.1190/1.1438976, doi:10.1190/tle37070529.1, doi:10.1190/segam2018-2997304.1, doi:10.1190/segam2018-2997865.1}. Applying such a trained network provides the probability that the central pixel of each patch is a fault/channel/salt/chimney/horizon or not. This strategy has the limitation that the network never sees the large-scale features present in large seismic images. Network architectures like autoencoders and U-nets \citep{Ronneberger2015} that map image-to-image do not have the same limitation, provided the data and label images are sufficiently large. These methods simultaneously classify all pixel in a large image, see \cite{wu2018deep, doi:10.1190/segam2018-2997085.1} for geoscientific applications. \cite{doi:10.1190/segam2018-2997085.1} notes that this advantage also creates a challenge when generating the labels. Whereas image-to-pixel networks can work with scattered annotations, standard network and loss functions for image-to-image classifications require the label images to be annotated fully, i.e., each pixel needs to have a known class.

We propose to use a partial loss function as introduced by \cite{SeisHorizons}, to be able to work with data and label images for which only a small subset of the pixels have known labels. This approach avoids the aforementioned difficulties when generating labels. We propose and investigate two strategies for choosing the pixels that need to be labeled, for the problem of segmenting a seismic image into distinct geological units. Our training data consists of a few dozen very large seismic images collected in a sedimentary geological area. We employ a U-net based network design that processes the seismic images on multiple resolutions simultaneously, therefore taking various length-scales of geological features into account explicitly. Numerical results evaluate the proposed labeling strategies. They show that if we take geophysical domain-specific information into account that is not available in other applications such as natural language processing or video segmentation for self-driving vehicles, we obtain better results using fewer labeled pixels. 

\section{Labeling strategies for seismic images}
Our goal is to develop strategies that require the least amount of time from a human interpreter of seismic images to generate the labels. Many test data sets for semantic segmentation (e.g., CamVid) come with fully annotated label images. Several geophysical works \citep{doi:10.1190/segam2018-2997085.1, di2018developing} also rely on full label images, which are time-consuming to create manually with high accuracy. Therefore, we propose two labeling strategies that require less user input. 

\begin{figure}[!htb]
   \centering
   \begin{subfigure}[b]{0.49\textwidth}
   	\centering
   	\includegraphics[width=1.0\textwidth]{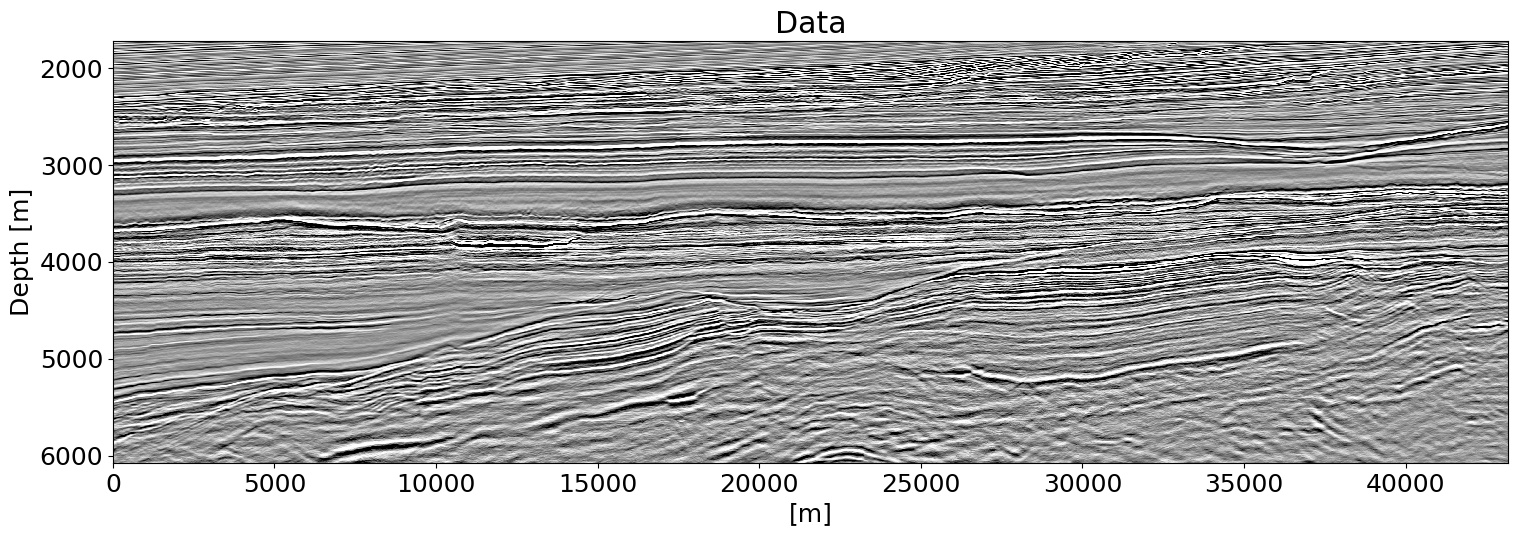}
   	\caption{}
   	\label{data_ex}
   \end{subfigure}
      \begin{subfigure}[b]{0.49\textwidth}
        \centering
   	\includegraphics[width=1.0\textwidth]{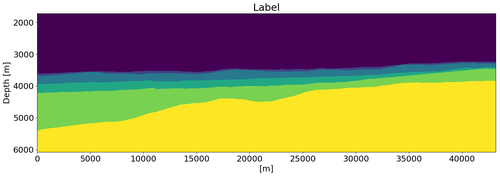}
   	\caption{}
   	\label{label_ex}
   \end{subfigure}
   \caption{(a) One of the 24 data images of size $1088 \times 2816$ pixels, and (b) corresponding full label (not used in this work).}
   \label{data_label_ex}
 \end{figure}

The interpretation information provided to the authors by an industrial partner comes in the form of x-y-z locations of a few important geological horizons. Because the geology is sedimentary, we assume that every horizon occurs at a single depth for a given x-y location. This information opens up the possibility to generate labels of geological units by assigning a class value to the area in between any to horizons. A full label image and the corresponding data are shown in Figure \ref{data_label_ex}. The number of occurances per class in the labels is evidently unbalanced.

\subsection{Scattered annotations}
Probably the simplest method to create label images is by selecting scattered points in the seismic image, and assign the class to just those pixels. If the intepreter investigates $n_\text{samp}$ locations in the seismic image, we obtain $n_\text{samp}$ label pixels. Questions regarding this strategy include 1) should we select an equal number of label pixels per class? 2) What if the interpreter only works on `easy' locations and does not create labels near the class boundaries? Answers are beyond the scope of this work, but in the numerical examples, we generate an equal number of label pixels per class, at randomly selected locations.

\subsection{Column-wise annotations}
Suppose the interpreter labels one column at a time. If $n_\text{horizons}$ horizons are marked in the $n_z \times n_x$ image, all points in between the horizon locations are also known. Labeling $n_\text{samp}$ pixels will thus result in $n_\text{samp} / n_\text{horizons} \times n_z$ labeled pixels. Provided that we are interested in a few horizons and geologic units, column-wise annotations yield a much larger number of label pixels per manual annotation point compared to scattered annotation. The second benefit is that column-based labeling samples points at, and close to, the class boundaries.

 \begin{figure}[!htb]
   \centering
   \begin{subfigure}[b]{0.49\textwidth}
   	\centering
   	\includegraphics[width=1.0\textwidth]{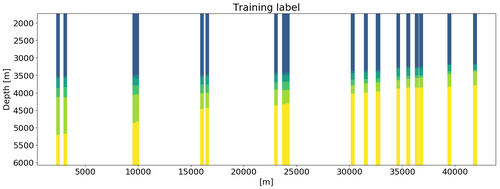}
   	\caption{}
   	\label{label_sub_column}
   \end{subfigure}
      \begin{subfigure}[b]{0.49\textwidth}
        \centering
   	\includegraphics[width=1.0\textwidth]{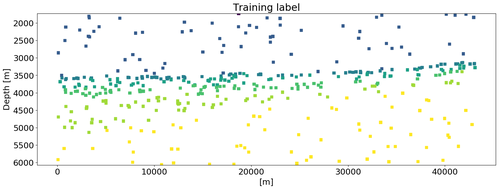}
   	\caption{}
   	\label{llabel_sub_rand}
   \end{subfigure}
   \caption{(a) A label image generated by column-wise annotations, and (b) randomly labeled pixels. Both labels are partial versions of Figure \ref{label_ex}. The colors indicate the class at each location, white space represents unknown labels that are not used to compute the loss or gradient.}
   \label{label_sub}
 \end{figure}
 
\section{Network and partial loss function}
Our network $f(\Ttheta,\by) : \mathbb{R}^N \rightarrow \mathbb{R}^{N \times n_\text{class}}$ is based on the U-net \citep{Ronneberger2015} and maps a vectorized seismic image, $\by \in \mathbb{R}^N$, of size $N = n_z \times n_x$ to $n_\text{class}$ probability images of the same size. The final interpretation follows as the class corresponding to the maximum probability per pixel. The network parameters $\Ttheta$ contain convolutional kernels and a final linear classifier matrix. We use the same network design as \cite{SeisHorizons}, which is a symmetric version of the U-net, with $37$ layers, and it has between $6$ and $32$ convolutional kernels of size $3 \times 3$ per layer. There are no fully connected layers, so that the network can use input-output of varying sizes.

Let us denote the number of training data/label examples (images) as $n_\text{ex}$. We represent the labels, the true probability per class per pixel, for a single image as $\TC \in \mathbb{R}_+^{N \times n_\text{class}}$. The networks in this work train by minimizing the cross-entropy loss, which for a single data image with corresponding full label reads

\begin{equation}
l(\by,\Ttheta,\TC) = - \sum_{i=1}^N  \sum_{j=1}^{n_\text{class}} \TC_{i,j} \log \bigg( \frac{\operatorname{exp} (f(\Ttheta,y)_{i,j})}{\sum_{j=1}^{n_\text{class}}\operatorname{exp}(f(\Ttheta,\by)_{i,j})} \bigg).
\end{equation}

The cross-entropy loss is separable with respect to the pixels (index $i$). To be able to work with partially known label images, we define the partial cross-entropy loss as

\begin{equation}\label{partial_ce}
l(\by,\Ttheta,\TC)_\Omega = - \sum_{i \in \Omega}  \sum_{j=1}^{n_\text{class}} \TC_{i,j} \log \bigg( \frac{\operatorname{exp} (f(\Ttheta,\by)_{i,j})}{\sum_{j=1}^{n_\text{class}}\operatorname{exp}(f(\Ttheta,\by)_{i,j})} \bigg).
\end{equation}
The set $\Omega$ contains all pixels for which we have label information. We thus compute the full forward-propagation through the network using the full data image, $f(\Ttheta,\by)$, but the loss and gradient are based on a subset of pixels only.

We train the network using Algorithm 1 from \cite{SeisHorizons}, which is stochastic gradient descent using one out of the $24$ data and label images per iteration. We use the partial cross-entropy loss instead of the partial $\ell_1$-loss because we classify each pixel for the segmentation problem. There are $120$ epochs, where we reduce the learning-rate by a factor $\times 10$ every $30$ epochs, starting at $1.0$.

\section{Results}

We evaluate the results of the labels with scattered and column-wise annotations. Our goal is a) to see which of the two sampling strategies performs best, given a fixed number of manually annotated pixels, and b) can we obtain highly accurate geologic classifications form a `reasonably' small number of labeled pixels? 

Both the seismic data and geologic interface locations from the Sea of Ireland were provided by an industrial partner. There are six classes, see Figure \ref{data_label_ex}. We give ourselves a budget of $100$ labeled pixels for each of the $24$ seismic images of size $1088 \times 2816$. We can thus choose to use label images with $100/5=20$ labeled columns because we need $5$ interface locations to obtain a fully labeled column, or $100$ randomly selected label pixels, i.e., $100/6 \approx 17$ per class. Figure \ref{label_sub} displays an example of a label for each labeling strategy.

 \begin{figure}[!htb]
   \centering
   \begin{subfigure}[b]{0.49\textwidth}
   	\centering
   	\includegraphics[width=1.0\textwidth]{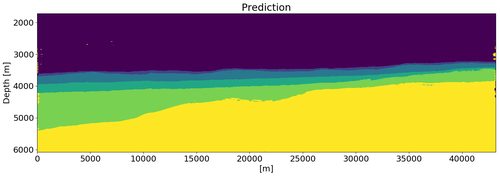}
   	\caption{}
   	\label{pred_thres_column}
   \end{subfigure}
      \begin{subfigure}[b]{0.49\textwidth}
        \centering
   	\includegraphics[width=1.0\textwidth]{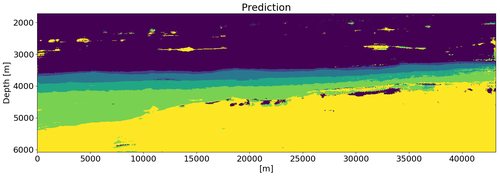}
   	\caption{}
   	\label{pred_thres_random}
   \end{subfigure}
         \begin{subfigure}[b]{0.49\textwidth}
        \centering
   	\includegraphics[width=1.0\textwidth]{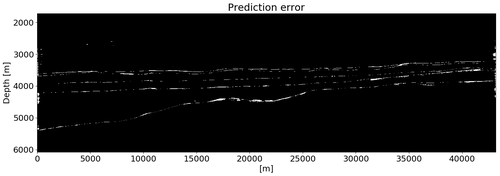}
   	\caption{}
   	\label{pred_thres_error_column}
   \end{subfigure}
         \begin{subfigure}[b]{0.49\textwidth}
        \centering
   	\includegraphics[width=1.0\textwidth]{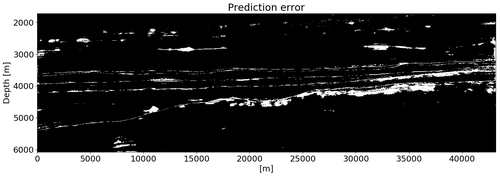}
   	\caption{}
   	\label{pred_thres_error_random}
   \end{subfigure}
   \caption{Classified seismic data using (a) column-sampling based training labels and (b) randomly sampled label pixels. Colors correspond to the class of the maximum predicted probability for each pixel. Errors in predicted class are shown in white in the bottom row, corresponding to the predictions above.}
   \label{preds}
 \end{figure}

The results in Figure \ref{preds} show that column-based annotations lead to more accurate predictions compared to generating labeled pixels at random locations. This verifies the, perhaps, expected result that a larger number of annotated label pixels provide more information. Additional experiments (not shown), reveal that more known label pixels lead to more accurate classifications. The maximum increase in prediction quality for the column-based labeling is limited, however, because the provided labels were generated manually by industrial seismic interpreters, and are not $100\%$ correct. Both labeling strategies achieve similar prediction accuracy when we increase the number of known label pixels to about $600$. We could also use data-augmentation to achieve higher prediction accuracy for a given number of labeled pixels, so the results serve as a baseline for the segmentation quality.

\section{Conclusions}
Interpreting seismic images by classifying them into distinct geologic units is a suitable task for neural-network-based computer vision systems. The networks map seismic images into an interpreted/segmented image. Networks that operate on multiple resolutions simultaneously can take both small and large scale geological structure into account. However, standard image-to-image networks and loss functions require target images for which each pixel has label information. This is a problem for geoscientific applications, as it is difficult and time-consuming to annotate large seismic images manually and completely. We presented a segmentation workflow that works with partially labeled target interpretations, thereby removing one of the main (perceived) barriers to the successful  application of neural networks to the automatic interpretation of large seismic images. We proposed and evaluated two strategies for generating partial label images efficiently. Generating column-wise labels is more efficient because a small number of interface annotations also provides us with all labels in between. The combination of a symmetric U-net, partial cross-entropy loss function, training on large seismic images without forming patches, and time-efficient labeling strategies form a powerful segmentation method.

\bibliographystyle{abbrvnat}
\bibliography{SeismicUnitClassificationPlain}{}

\end{document}